\documentclass{IEEEtran} 
\usepackage{times}  
\usepackage{url}
\usepackage[hidelinks]{hyperref} 
\usepackage[T1]{fontenc} 

\usepackage[small]{caption}
\usepackage{graphicx}
\usepackage{amsmath}
\usepackage{booktabs} 
\usepackage{algorithm}
\usepackage{algpseudocode}

\usepackage{color}
\usepackage{multirow}
\usepackage{xr}
\usepackage{listings}

 \renewcommand{\diamond}{{\color{red}\diamondsuit\color{black}}}
 \newcommand{\heart}{{\color{red}\heartsuit\color{black}}}
 \newcommand{\club}{{\color{black}\clubsuit\color{black}}}
 \newcommand{\spade}{{\color{black}\spadesuit\color{black}}}

\title{Knowledge-Based Paranoia Search in Trick-Taking}

\author{
Stefan Edelkamp, AI Center at CTU Prague, Czech Republic
}

\begin{document}

\maketitle

\begin{abstract}
This paper proposes \emph{knowledge-based paraonoia search} (KBPS) to find forced wins  during trick-taking in the card game Skat; for some one of the most interesting card games for three players.  It combines efficient partial information game-tree search with knowledge representation and reasoning. This worst-case analysis, initiated after a small number of tricks, leads to a prioritized choice of cards.
We provide variants of KBPS for the declarer and the opponents, and an approximation to find a forced win against 
most worlds in the belief space. 
Replaying thousands of expert games, our evaluation indicates that the AIs with the new algorithms perform better than humans in their play, achieving an average score of over 1,000 points in the agreed standard for evaluating Skat tournaments, the extended Seeger system. 
\end{abstract}

\lstset{language=C++,morekeywords={array,constraint,var,forall,sum,solve,minmize,decreasing,domain},basicstyle=\ttfamily,keywordstyle=\color{blue}\ttfamily,
literate=%
  {+}{{{\color{red}+}}}1
  {!}{{{\color{red}!}}}1
  {*}{{{\color{red}*}}}1
  {/}{{{\color{red}/}}}1
  {=}{{{\color{red}=}}}1
  {|}{{{\color{red}|}}}1
  {\%}{{{\color{red}$\%$}}}1
  {<}{{{\color{red}<}}}1
  {~}{{{\color{red}$\sim$}}}1
  {\&}{{{\color{red}\&}}}1 
  }

\newtheorem{definition}{Definition}
\newtheorem{theorem}{Theorem}
  \setlength{\tabcolsep}{5pt}

\section{Introduction}

One central showcase of artificial intelligence is to prove that computers are able to beat humans in games~\cite{SCHAEFFER2000189}. 
Success stories in playing games have been highly influential for AI research in general~\cite{AlphaGo}.

As many board games have either been solved~\cite{checkers} or AIs show superhuman performance~\cite{alphazero}, one of the next AI challenges are card games with randomness in the 
deal and incomplete information due to cards being hidden. 
While there is impressive research on playing non-cooperative card games like Poker~\cite{poker}, for multi-player trick-taking games, human play appears better to computer play. 
Despite early ground-breaking results, e.g., in Bridge~\cite{GIB}, according to~\cite{DBLP:journals/corr/abs-1911-07960} best computer Bridge programs still play inferior to humans. 

Another candidate for showing the intriguing challenges in trick-taking card play is Skat, an internationally played game, described by McLeod as the best and most interesting card game for three players\footnote{\url{www.pagat.com}}.  
Skat has a deck of 32 cards; in a deal each player gets 10 cards, with two left-over Skat cards.
There are four stages of the Skat game: $i)$ \emph{bidding}, where the players communicate values towards their respective maximal bidding strength; $ii)$ \emph{Skat taking} and selecting the game; $iii)$ choosing the two cards for \emph{Skat putting};  $iv)$ \emph{trick-taking game play} with up to 10 rounds of play. For higher bidding values, stages $ii)$ and  $iii)$ may be skipped.

There are three main games types played in Skat: \emph{Grand}, where only Jacks are trump, \emph{Suit}, where Jacks and the selected suit is trump, and \emph{Null}, a variant, where all tricks must be lost. 
The winner of the bidding becomes the declarer, who plays against the remaining two opponents. S/he adds the Skat to his/her hand and discard any two cards. 
The declarer wins if s/he gets more than 60 points (Grand or Suit) or makes no tricks (Null).
To increase the bidding value further, there s/he can raise the contract 
from scoring 61 points to 90 (\emph{Schneider}) and 120 (\emph{Schwarz}), and also to open the hand to the opponents (Ouvert). 
%
Handling partial information is the critical aspect in this game, given that for open card play, the optimal score and the associated playing card can be found in terms of milliseconds~\cite{kupferschmid:masters-thesis-03}.


The contribution of this paper is \emph{knowledge-based paranoia search} (KBPS). The widely applied Perfect-Information Monte-Carlo Sampling (PIMC)~\cite{DBLP:conf/aaai/SolinasRB19,DBLP:journals/corr/abs-1911-07960} may not find a paranoid strategy, even if one exists. Due to the problem of strategy fusion, it might return a move corresponding to lines of play that lack knowing the true world.
Every time a player is
to act, the legal moves are all possible moves over the current set of consistent worlds. Whenever
a player makes a move, we restrict the consistent worlds to only those worlds where the given
move was actually legal~\cite{skatfurtak}. In case of KBPS, we 
include knowledge representation and reasoning into the search and 
run the analysis after the third trick for each card to be played.

We include the proposal into a Skat AI, able to play all stages of the game and all game types. Using statistical tables elicited from Human expert games, it derives accurate winning probabilities, which are used mainly for the bidding and game selection stages, and to put good Skats. For the trick-taking stage of the game, it includes algorithmic strategies for opening, middle- and endgame play using expert rules and exploiting playing conventions to build a knowledge base on plausible and effective locations of the cards.
For the opening stage, using winning features statistics from expert games are stored in pre-computed tables~\cite{DBLP:conf/socs/Edelkamp19}. For the middle game we apply {suit-factor search}~\cite{DBLP:conf/ki/Edelkamp20}.
For the endgame the remaining space of possible beliefs is analyzed completely and the strategies in terms of recommended cards are fused using a voting scheme~\cite{DBLP:conf/ecai/Edelkamp20}.

\section{About Skat}

Skat is a three-player imperfect information game played with 32 cards, a subset of the usual 52 cards Bridge deck.
It shares similarities to Marias(ch) (played in Czech Republic and Slovakia)
and Ulti (played in Hungary).

\begin{figure}
    \centering
    \includegraphics[width=9cm]{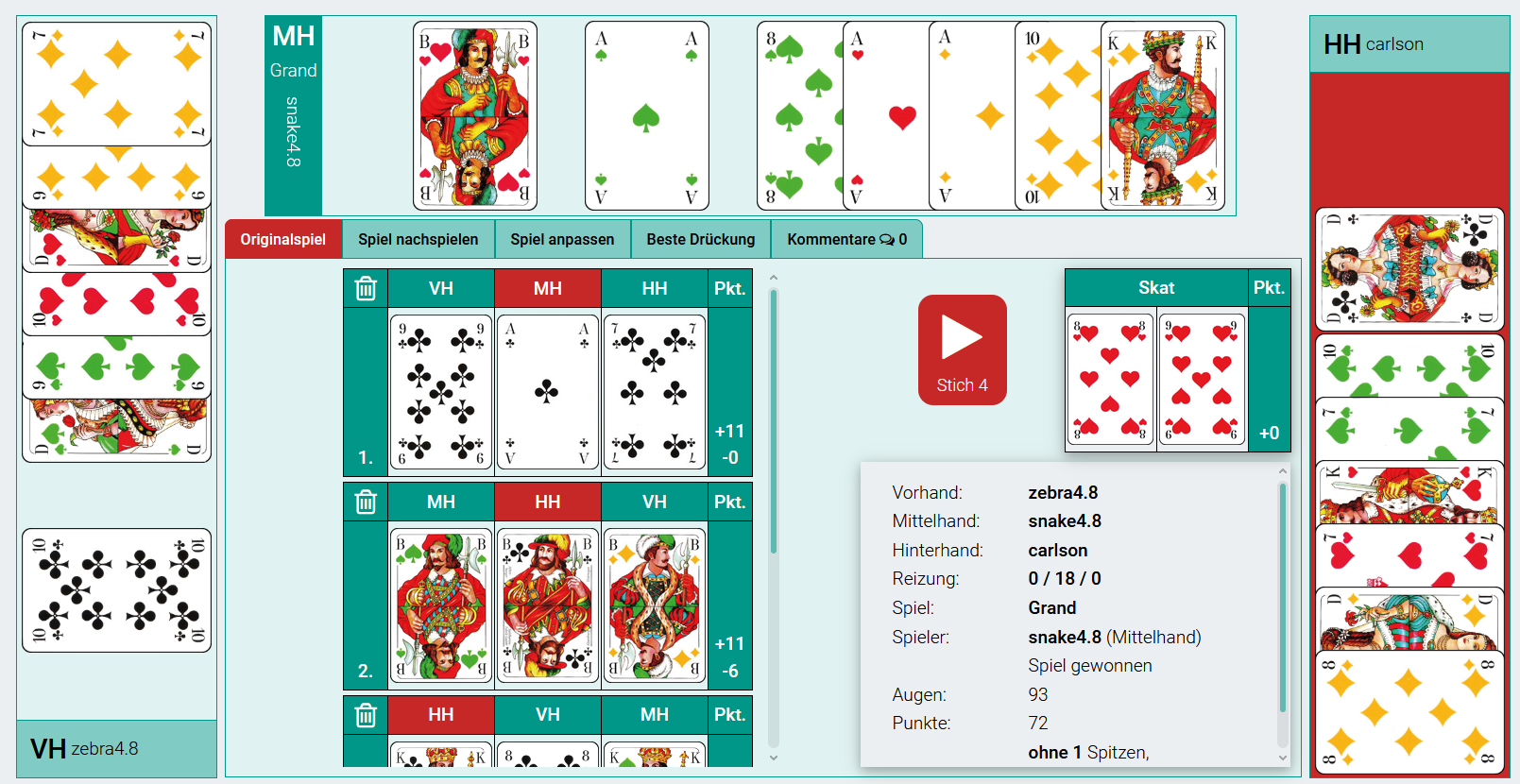}
    \caption{Skat game replay of our AI playing against a human and another AI. }
    \label{fig:skatput}
\end{figure}

At the beginning of a game, each player gets 10 cards, which are hidden to the other players. The remaining two cards, called the Skat, are placed face down on the table. Each hand is played in two stages, bidding and card play. 

The bidding stage determines the declarer and the two opponents: two players announce and accept increasing bids until one passes. 
The winner of the ﬁrst bidding phase continues bidding with the third player. 
The successful bidder of the second bidding phase plays against the other two. The maximum bid a player can announce depends on the type of game the player wants to play and, in case of a trump game, a multiplication factor determined by the jacks. The solist decides on the game to be played. Before declaring, he may pick up the Skat and then discard any two cards from his hand, face down. 
These cards count towards the declarer's score. 
An example for Skat selection is shown in Fig~\ref{fig:skatput}.

Card play proceeds as in Bridge, except that the trumps and card ranks are different. In grand, the four jacks are the only trumps. In suit, additional seven other cards of the selected suit are trumps. There are no trumps in null games. Non-trump cards are grouped into suits as in Bridge. Each card has an associated point value between 0 and 11, and in a standard trump game the declarer must score more points than the opponents to win. Null games are an exception, where the declarer wins only if he scores no trick.

\section{Related Work}
 
The game of Skat is considered in many books~\cite{Lasker2,Wergin,Grandmontagne,Kinback,Quambush,Harmel}. A recent mathematical introduction to Skat playing has been given by~\cite{Rainer}. There are frequent bachelor and 
master theses on the topic of Skat (e.g., by 
Fabian Knorr, 2018, University Passau, or by Dennis Bartschat,
2019, University of Koblenz), but due to the limited time for programming, the proposed Skat bots do not reach a human-adequate playing strength.

Kupferschmid and Helmert \cite{doubledummy} developed the \emph{double-dummy Skat solver} (DDSS), a fast open card Skat game solver, which found its way into the
\emph{Kermit} player~\cite{DBLP:conf/ijcai/BuroLFS09}. DDSS was
extended to cover the partial observable game using Monte-Carlo sampling~\cite{GIB}. It reached only moderate performance results, mainly due to lacking knowledge information exchange between the players.

There have been larger efforts to apply machine learning to predict bidding options and hand cards in Skat~\cite{biddingskat,DBLP:conf/ijcai/BuroLFS09,DBLP:journals/corr/abs-1903-09604,DBLP:journals/corr/abs-1905-10907,DBLP:journals/corr/abs-1905-10911,skatfurtak}. Additionally, we have seen feature extraction in the related game of \emph{Hearts}~\cite{DBLP:conf/cg/SturtevantW06}, and automated
bidding improvements in the game of \emph{Spades}~\cite{DBLP:conf/ecai/CohensiusMOS20}. 
The results show that the prediction accuracy can be improved.

Buro et al.~\cite{DBLP:conf/ijcai/BuroLFS09}
indicate that their player Kermit achieved
expert-playing strength. A direct comparison 
is rather difficult, as the 
current bots play on different server architectures and use different communication protocols.

Cohensius et al. \cite{DBLP:conf/ecai/CohensiusMOS20}
elaborated on
an intuitive way of statistical sampling 
the belief space of hands (worlds) based on the knowledge inferred within play. The matrices $P^i$ for the belief of card location for each player $i$ show a probability $p^i_{j,k}$ for the other players $j$ on having a card $k$ in his/her hand. While the approach has been developed for \emph{Spades}, it also applies to 
Skat~\cite{DBLP:conf/aaai/SolinasRB19}. 

In a different line of research, Edelkamp~\cite{DBLP:conf/socs/Edelkamp19}
showed how to predict winning probabilities for the early stages of the game, and how to play the Null game. Edelkamp \cite{DBLP:conf/ecai/Edelkamp20})
 studied Skat endgame play using a complete analysis of the belief-space that is compactly kept and updated in knowledge vectors. Referring to 
combinatorial game theory~\cite{Conway}, Edelkamp
\cite{DBLP:conf/ki/Edelkamp20}
proposes suit factorization and mini-game search for improved middlegame play in Skat.

Sturtevant and Korf \cite{para} described a paranoid algorithm for the case
of perfect information multi-player games. In that work the player to act is paranoid with respect
to the preferences of the other players, assuming that they are in a coalition against the agent. This
reduces the multi-player game to a 2-player game, such that $\alpha$-$\beta$ pruning may be applied.
Partial-information $\alpha$-$\beta$ 
paranoid search has been considered by Furtak \cite{skatfurtak}, 
The work differs from Sturtevant and Korf’s algorithm in that the agent does not have perfect
information. Moreover, because the agent does not know the true world, it is also paranoid wrt.\ the outcome of any stochastic events (chance nodes), namely the actual distribution of
any unobserved cards. The information was used for bidding and stored in tables for the $(32!/ 10!2!) \cdot 3 \cdot 5 \approx 224$ million hands (including game type and
turn), symmetry-reduced and compressed.

Edelkamp \cite{DBLP:conf/ecai/Edelkamp20} considered a similar paranoia partial information search option for analyzing Skat puzzles mainly 
as a motivating aspect to introduce knowledge representation and reasoning in bitvector sets for endgame play. As with 
Furtak \cite{skatfurtak}, the algorithm never went into the players' trick-taking stage, as it was cast inefficient to be useful under real-world playing constraints. 
In this work, we successfully integrate this analysis option into actual game play, leading to a considerable increase in playing strength.

\section{Architecture of Skat AI}

Our Skat AI is an implementation on top of the player of Edelkamp \cite{DBLP:conf/ecai/Edelkamp20,DBLP:conf/ki/Edelkamp20}, kindly provided by the author. 

Bidding and game selection both use statistical
knowledge of winning ratios in human expert games, 
stored in tables and addressed via 
patterns of \emph{winning features}.
This assumes a predictor 
for a given hand with high accuracy, before 
play (no move history). 
We assume the player's position
to be part of his/her hand. 
The winning probability of a hand decreases during during by the anticipated larger strength of the opponent
hand. Other than this, so far no opponent 
model is used. 

The Skat bot estimates winning probabilities with statistical tables that are extracted from a database of millions of high-quality expert games; more precisely, winning probabilities $Prob(h,p,s,b,t)$ including current hand $h$, choices of game type $t$, Skats $s$, position
of the player $p$, and bidding value $b$. The probabilities are then used in the first three stages of the Skat game: bidding, game selection, and Skat putting. For each bidding value and each game type selected, it generates and filters all ${22 \choose 2}=231$ possible Skats and takes the average of the payoff of skat putting, which, in turn, is the maximum of the ${12 \choose 2} = 66$ possible skats to be put. 
The winning ratios in expert games can easily be analyzed statistically, but by the high number of 
$n={32 \choose 10}{22\choose 10}{12 \choose 10} \approx 2.8$ quadrillion possible Skat deals, proper generalizations are needed. 

For Null games, given hand $h$ and skat $s$ the approach estimates the winning probability $Prob(h,s) = Prob(h,s,\club) \cdot Prob(h,s,\spade) \cdot Prob(h,s, \heart  ) \cdot Prob(h,s,\diamond)$~\cite{Lasker,Lasker2}.
For trump games, we consult a table addressed by the so-called winning parameters: number of non-trump suits that the player lacks;
number of eyes in Skat, condensed into four groups;
value of the bidding stage, projected to four groups;
position of the declarer in the first trick;
number of trump cards in hand;
number of non-trump cards in hand;
constellations of jacks condensed into groups; and number of cards estimated to lose, based on summing the expected number of standing cards.
Statistical tests~\cite{Rainer} showed that these parameters have a significant influence and can, therefore, be used as essential attributes to accurately assess the probability of winning a trump game. In particular, a \emph{Grand} table with $113,066$ entries is built on top of 7 of these winning parameters and a \emph{Suit} table with $246,822$ entries using 9 of them. For Skat putting we refine the lookup value for different cases in a linear function together with further winning features such as the expected number of tricks while respecting the retaking options of the issuing right, and the exact number of points put into the Skat.

Trick-taking is arranged wrt.\ an ensemble of different card recommendations. For the sake of brevity, we refer reader to precursor work~\cite{DBLP:conf/socs/Edelkamp19,DBLP:conf/ecai/Edelkamp20,DBLP:conf/ki/Edelkamp20}. In short terms, we find 
\begin{itemize}
    \item killer cards that force a win for the declarer (or the opponents) to meet (or to break) the contract of the game; this option mainly includes the KBPS card proposals of this paper; other are simpler rules that count the number of points certain to be made for the player to move in the remaining tricks.
    \item endgame cards as the results of strategy fusion, realized via a voting on the winning ratio of open card game solver calls on the remaining \emph{worlds} in the belief space of the player~\cite{DBLP:conf/ecai/Edelkamp20}. 
    The endgame player is invoked after five tricks with a maximum number of $2500$ worlds in the belief space, 
    the win ratio for a card (confidence level) is set to 
    $90\%$. Additional bonus is given for a high number of eyes and for meeting higher contracts.
    \item hope cards as the only moves that can safe the game for either the declarer or the opponents, i.e., all other cards lead to a forced loss, this card is played instantly
    \item expert cards for each player in each position in the trick, based on if-then-else rules that consider the current the hand of issuing players, the history of tricks being played, the partial knowledge of cards present in the opponent hands, etc.
\end{itemize}
 
The priority is as follows. First, killer card are recommended; if this strategy fails to find a forced win, endgame and hope cards are searched for; if this does not meet the required criteria or confidence level, we fall back to expert cards recommendation. The expert rules, used for the first few tricks and as a default, includes card recommendations based on suit factors (either trump or non-trump). Each card in the factor is assigned a value $0$, $1$, or $2$, where $0$ denotes a hand card, 
$1$ a card in the other players' hands,
and $2$  a card that is not playable (either being played or put into the Skat). For the declarer issuing trump we precomputed tables of sizes
${11 \choose k}\cdot 2^{11-k}=$
$11,264$ ($k=1$ trump),
$28,160$ ($2$ trumps), 
$42,240$ ($3$ trumps),
$42,240$ ($4$ trumps),
$29,568$ ($5$ trumps),
$14,784$ ($6$ trumps),
$52,80$ ($7$ trumps),
$1,320$ ($8$ trumps),
$220$ ($9$ trumps), and
$22$ ($10$ trumps).
For non-trump suits we
generated a table with $\sum_{k=1}^7 {7 \choose k}\cdot 2^{7-k}=3^7-2^7=2,059$ entries.

\section{Knowledge Representation and Reasoning}

We represent the knowledge in the players as bitvector sets.
To introduce the reasoning on the sets we give a brief example.

Suppose we have the following deal with declarer $P_0$ 
and opponent player $P_2$ to issue the first card: \\
 $Player_0: \heart    J, \diamond   J, \heart A, \heart  K, \heart9, \heart7, \club A, \club8, \club7, \spade A,$ \\
 $Player_1: \club  J, \spade  J, \heart Q, \club T, \club K, \club Q, \spade T, \spade7, \diamond Q, \diamond 7$ \\
 $Player_2: \heart T, \heart8, \club9, \spade K, \spade Q,  \spade9, \spade 8, \diamond A, \diamond T, \diamond8$ \\
 Skat: $\diamond K, \diamond9$ \\
 The game that is being played is $\heart$.    
 
In set representation we have the following initial knowledge for $P_2$: \\
$P_0 = \{\}$ \\
$P_1 = \{\}$  \\
$P_2 = \{\heart T, \heart8, \club9, \spade K, \spade Q, \spade 9, \spade 8, \diamond A, \diamond T, \diamond8\}$ \\
$pool = \{\club  J, \spade  J, \heart   J, \diamond  J, \heart A, \heart K, \heart Q, \heart9, \heart7, \club A, \club  T,$ $ \club K, \club Q, \club8, \club7, \spade A, \spade T, \spade7, \diamond K, \diamond Q, \diamond9, \diamond7\}$ \\
$skat  = \{\}$ \\ 
$declarerorskat = \{\}$ \\
$partnerorskat = \{\}$ \\
$noskat = \{\club  J, \spade J, \heart J, \diamond J, \heart A, \heart T, \heart K, \heart Q, \heart 9, \heart 8, \heart 7$, $\club A, \spade A, \diamond A \}$ \\
The declarer then sees $\diamond A$ on the table and updates his/her knowledge\\
$P_0 = \{\heart J, \diamond  J, \heart A, \heart K, \heart9, \heart7, \club A, \club8, \club7, \spade A\}$ \\
$P_1 = \{\}$ \\
$P_2 = \{\}$  \\
$skat  = \{\diamond K, \diamond 9\}$ \\ 
$pool = \{\club  J, \spade  J, \heart Q, \heart T, \heart8, \club  T, \club K, \club Q, \club9, \spade T$, $\spade K, \spade Q, \spade9, \spade8, \spade7, \diamond T, \diamond Q, \diamond8, \diamond7\}$ \\
Player 1 now sees $\diamond A, \heart A$ on the table and
updates his/her knowledge\\
$P_0 = \{\}$ \\
$P_1 = \{\club  J, \spade  J, \heart Q, \club  T, \club K, \club Q, \spade T, \spade7, \diamond Q, \diamond 7\}$ \\
$P_2 = \{\}$  \\
$skat  = \{\}$ \\ 
$pool = \{\heart   J, \diamond  J, \heart  K, \heart T, \heart9, \heart8, \heart7, \club A, \club9, \club8, \club7$, $\spade A, \spade K, \spade Q, \spade9, \spade8\}$ \\ 
$declarerorskat = \{\}$ \\
$partnerorskat = \{\diamond T, \diamond K, \diamond9, \diamond8\}$ \\
$noskat = \{ \club  J, \spade  J, \heart   J, \diamond  J, \heart T, \heart  K, \heart Q, \heart9, \heart8, \heart7$, $\club A, \spade A\}$ \\

Given the sets it is an easy task to convert the 
representation into a matrix of probabilities.

At the moment we have two times four reasoning algorithms
\emph{ObeySuit},
\emph{DropOnSuit},
\emph{ObeyTrump}, and
\emph{DropOnTrump}
that are called in the accoding setting in middlehand or rearhand.

\section{Knowledge-based Paranoia Search}

The general approach is to solve a card game with randomness in the deal and partial information is to compute approximate Nash equilibria e.g., using counterfactual regret minimization~\cite{poker}. As this this
isn't feasible, approximations have to be found - of which inference, sampling, paranoid search, etc. are some examples.
We identified two different approaches for the search of playing cards with uncertainty. One is to generate a set of possible (or all) worlds coherent with the generated knowledge, and, then, to merge the result, possibly improved with dominance checks~\cite{DBLP:journals/corr/abs-1911-07960}. This is what is done during endgame play~\cite{DBLP:conf/ecai/Edelkamp20}. When the set of worlds is statistically Monte-Carlo sampled wrt.\ the knowledge of the distribution, as described in Jeff Rollason's article for the AI factory\footnote{\footnotesize \url{www.aifactory.co.uk/newsletter/2018_02_opponent_hand.htm}}, 
bias can be given to the distribution.
However, the approach often misses the best playing card in early stages of the game, when less knowledge is available.

The number of declarer cards unknown to
the opponents is important for many moves. Is it only one card, the
is no need to cut low, as the card will likely be put into the Skat. If the
opponents issues, he might use a 
\emph{sharp 10} to win the game.
One can also determine if a victory
can be enforced or Schneider avoided.

 Timothy Furtak's Ph.D. thesis \emph{Symmetries and
Search in Trick-Taking Card Games}, first 
describes a fast paranoid
search for creating paranoid hand databases~\cite{skatfurtak}.
Edelkamp \cite{DBLP:conf/ecai/Edelkamp20}  also presented a first attempt for conducting a search for a forced win
against all odds, aimed at the first card of the Skat game. While interesting for solving Skat puzzles in newspapers, the running time for the analysis, however, was way too large to assist actual play given the restrictions to select a card imposed by the play clock (on our server a timeout is set to 5 seconds).

To alleviate the computational burden, we propose 
the search to be initiated only after a few tricks have been played. The algorithm has been adapted
to the knowledge already inferred by the Skat AI. It,
thus, takes as an input \emph{knowledge sets}~\cite{DBLP:conf/ecai/Edelkamp20}
corresponding to the inference that player $P_i$ must have cards $C_j$ (not) in his/her hand. This knowledge is inferred e.g., by unrealistic bad skats, players not obeying trump or non-trump cards and by playing conventions (putting the lowest-valued card in the declarers trick, and the highest-value trick to the one of my partner, with some exceptions).
As with many other parts of the Skat AI,
for efficiency reasons, sets of cards are encoded as 
bit-vectors of length 32 (unsigned int). This allows
fast bit manipulation, such as card selection and copying.

The minimax alpha-beta simulating \emph{moving test-driver} search algorithm~\cite{kupferschmid:masters-thesis-03} 
to analyze partial information trump games is implemented as a
binary search over an AND/OR tree decision procedure
that returns, whether or not the declarer can win the game according to a given
a contract limit. 
It progresses belief sets for partial information. The knowledge-based Paranoia search (KBPS) algorithms are applied in forehand, in middlehand, and in rearhand positions of the players. Besides updating knowledge vectors, scoring values, current contract limit, the call has to respect played cards on the table to trigger a correct analysis\footnote{By tracking server logs, we monitored the algorithm: once a win is established this determines the outcome of the game, in many cases before the human opponent 
recognized that s/he is lost.}.

\subsection{Paranoia Search for the Declarer}

The KBPS worst-case analysis for the declarer is used in 
trump games. Its implementation is a loop over backtracking moving test driver branch-and-bound procedure to find the optimal game value. As we use the search option dynamically, the alorithm is initiated after a fixed number $k$ of played cards. In the overall architecture it acts as prioritized killer card proposal that warrants a forced win.

The function \emph{paranoia} search takes the
partially played game, and a
set of possible worlds as a parameter, 
encoded as knowledge (bitvector) sets,
and contract bound. We limit the uncertain knowledge 
to the sets of \emph{free} cards that are still
to be distributed among the two opponent players. 
All \emph{fixed} cards are assigned to one hand.

Figure~\ref{double} shows the implementation of the declarers' KBPS backtracking algorithm at an OR node for the first opponent in the AND-OR partial observable search tree. 
It determines if a game can be won against all world according to a given score bound $limit$ as fixed by the overall binary search.
For the sake of simplicity, we omit code for transposition table pruning~\cite{DBLP:journals/pami/ReinefeldM94} and for pruning of equivalent cards~\cite{GIB}.
As the declarer knows that Skat (except for \emph{Hand} games), 
as with the above overall knowledge representation and reasoning 
example there are 3 knowledge sets provided to the player: $pool$, denoting all remaining cards not yet known on which opponent hand they reside,
$h_1$, cards already known to be in the 1st opponent hand, 
$h_2$, cards already known to be in the 2nd opponent hand.
Furthermore, we have  $avail$: hand
cards playable according to the rules of Skat, obeying trump and
suit; 
$index$, $bit$: selected card,
for being played; $played$: cards already played;
$w$: winner of trick;
$i_0,\ldots,i_2$: table cards by players;  
$r_1$, $r_2$ number remaining cards available;
$limit$: current bound for game value; 
$score$: card value of table cards;
$aspts$: points accumulated for the declarer (according to the given knowledge of the Skat);
$gspts$; points accumulated for the opponents.

The KBPS algorithm searches the tree of all playable cards, and branches according to the current set of known cards and
the current belief respecting the rules of play, and the number of cards that a player can have. If suits are not obeyed, knowledge vectors for cards available to each hand are updated during the search. Before cards are selected from the pool of cards available to both players they are assigned to one opponents' hand.

\begin{figure}[t]
\begin{scriptsize}
\begin{lstlisting}
OR1(avail)
  avail = playable(avail,h[1],i)  
  while (avail) 
    index = select(avail); bit = (1<<index);
    h2 = h[2]; o = pool;
    if (c = firstcardontable(i)) 
      if (trump & (1 << c))
        if (|trump & bit| == 0) 
          h[2] |= trump & pool; 
          if (|h[2]| > r2 - rearhand) 
            h[2] = h2; avail &= ~bit; continue;
          pool &= ~h[2];
     else 
       if (|suit(c) & bit| == 0)
          h[2] |= suit(c) & pool;
          if (|h[2]| > r2 - rearhand 
            h[2] = h2; avail &= ~bit; continue;
          pool &= ~h[2];
    if (|h[1]|bit| > r1) 
      (h[2],pool) = (h2,f); avail &= ~bit; continue;
    (h1,p,i[1],r) = (h[1],played,index,-1);
    pool &= ~bit; h[1] &= ~bit; played |= bit;
    if (endoftrick(i))
      r1--; r2--; w = winner(2,0,1);
      score = value(i);
      gspts += w ? score ; 0; aspts += w ? 0 : score;
      (i0,i2) = (i[0],i[2]); i[0] = i[1] = i[2] = -1;
      r = (gspts >= 120-limit) ? 0 : (aspts > limit) ? 1 :
        (w == 0) ? AND(h[0]) : 
        (w == 1) ? OR1((pool|h[1]) & ~h[2]) :
        OR2((pool|h[2]) & ~h[1]);
      i[0] = i0; i[2] = i2;
      gspts -= w ? score : 0; aspts -= w ? 0 : score; 
      r1++; r2++;
    else r = OR2((pool|hands[2]) & ~hands[1]);
    (i[1],h[1],h[2],pool,played) = (-1,h1,h2,o,p);
    avail &= ~bit;
    if (r == 0) return 0;
  return 1;
\end{lstlisting}
\end{scriptsize}
\caption{KBPS for the declarer at an opponent search node;
}\label{double}
\end{figure}

The algorithm can extended to cover more knowledge inference options like playing conventions for the opponents such as
giving the highest-valued card to a trick that goes to the partner, 
and a lowes-valued card to a trick the declarer.

If the capacity of a hand is exceeded, we encountered a deadend and a backtrack is initated. In other words, if more cards are assigned to the player than his/her hand can hold, the entire subtree is pruned. By the virtue of enumeration of all card combinations, the algorithm computes the game-theoretical partial information (minimax) score, assuming optimal play of the players. The transposition table and equivalent card pruning are implemented in a way not to violate this outcome.

A proof that a win is forced and will not be lost during subsequent play may be done by induction on the number of
remaining cards to be played, but is quite obvious, as the
game-theoretical minimax value is computed at the root node. 
As long as the knowledge is exact, i.e., given that no false information is contained in the knowledge vectors, then the 
algorithm progressing the vector does not falsify it. 

One argument is as follows. 
If there is only one card left for the declarer, the score of 
the last trick  determines the outcome of the game. 
Inductively, we see that given the same contract limit,
the AND/OR search tree for the 
declarers' choice in trick $t$, is part of the search tree for the declarers' choice in trick $t+1$. 
The search tree is spanned by a) the valid card choices of the players obeying trump and non-trump and
b) the uncertainty on where the opponent cards are. During the search, the free cards are distributed to either hand, as far as the number of known cards for a hand do not exceed the maximum possible. 

The knowledge only increases over time and is assumed to be exact. Hence, if the number of remaining
cards increases the game that could not be proven lost, cannot
be proven a win earlier on. Vice versa, depending on the 
card choices of the
opponents being optimal, the optimized point value can only
go up from trick to trick, so if a win is established for the declarer in trick $t$ will persist in trick $t+1$ 
during subsequent play. 

\begin{theorem} [Soundness KBPS for Declarer Play]
Assume that accurate knowledge is given to the KBPS algorithm in form of knowledge vectors, spanning the space of possible beliefs. If a win for the declarer has been found by the KBPS algorithm, then the
game is won and this win will manifest during subsequent
trick-taking play.
\end{theorem}

We are not claiming correctness, 
the opposite must not hold: if no win is found, then the
game is not necessarily lost but continues with other card recommendations.

\subsection{Paranoia Search for Schneider \& Schwarz}

When a game can be won to the contract of 61 points, it is desirable
to aim at Schneider (90 pts) or Schwarz (120 pts). This is done
by restarting the analysis with a higher contract, once the one for the current limit has been proven to be a win.

\subsection{Approximate Paranoia Search}

The worst-case analysis has two major limitations. 

\begin{enumerate}
\item As stated in Theorem~1, the AIs act \emph{in paranoia}. Suppose that all non-trump card of a suit are neither in the declarers hand nor in the Skat, then even extreme
distributions of the cards with all cards on either hand
have to handled in the analysis. The probability for this to happen is only $2/2^7=1.5625\%$. The virtue of good Skat play, however, is to play well against most likely card distributions.
For the approximate KBPS algorithm we, therefore, demand
that certain distributions of cards are unlikely and are excluded from
the search. 
\item 
Secondly, the running time is larger in case of more 
uncertainty, so that belief space
measured in the number
of worlds the AI plays against, is maybe to finish a complete
KBPS exploration in time.
\end{enumerate}

Both objections can be met together by limiting the 
cards that can be assigned to each hand. This is
the basis of the approximate knoledge-basesd paranoia search (AKBPS), that poses constraints on the cards distributions allowed on each hand,
or ---for implementation purposes--- enforces some 
cards assigned to a hand. Of course, the theorem no
longer holds, as there are some worlds that are not considered, still
the observation is that an early suggestion of a card that wins 
against all but extreme worlds is valuable. 

In contrast, Furtak used lower bounds derived from Paranoia search, and, because of pessimistic assumptions sets the cut-off for the declarer to 57.

\subsection{Paranoia Search for the Opponents}

Extending the approach from the declarers' point of view to the ones of the opponents is
tricky, mainly due to the presence of the unknown Skat. 
 
For example, if one of the other players does not obey, it is no longer immediate that the card is on the remaining player's hand, as it can reside in the Skat. In the knowledge-based paranoia search algorithm, illustrated for the case of the declarer's AND node in Figure~\ref{triple} (for efficiency reasons, we 
are using many bitvector set operations!)
this leads to the introduction of further knowledge vectors. 
We now have five sets that are updated denoting that the declarer or the Skat has a card, or that the opponent, or the Skat has a card of the pool of remaining cards, only if taken or a card is definitely known to be on the hand, e.g., by selecting it, it is moved. In some respect the knowledge sets ($declarerorskat$ and $partnerorskat$) are caches for the main pool of cards  ($pool$) for the remaining players. In some of the conditions applied we take care that no more cards are moved to a hand that it can cope. 

Note that if one opponent sees a definite win, this does
not mean that the other opponent sees it as well. Given 
a different set of hand cards s/he may have very
different knowledge on the distribution of cards.
As it is defined, it requires one defender to assume that his partner will intentionally play poorly. 

Again, a soundness proof can be found by induction of the
remaining cards to be played, and the observation that a search
tree with less remaining cards is part of a search tree with more
remaining cards, leading to a forced win. 
According to the uncertainty in the Skat there are three pools of cards that reflect the rising knowledge instead of one. 
Cards in the general pool of cards  move to declarer-or-Skat, or to the opponent team player.

\begin{theorem} [Soundness KBPS for Opponent Play]
Given that the knowledge is accurate, 
if a win has been found by the KBPS algorithm, the
game is a forced win by the opponent to move and the win will manifest during subsequent trick-taking play.
\end{theorem}

Again, we are not claiming correctness of the algorithm.

\subsection{Worst-Case Analysis for Avoiding Schneider/Schwarz}

In opponent play, using a paranoid assumption on 
the card play is less effective than for the declarer
play, and often applies to the endgame analysis. 
When the game is won by the declarer, however, KBPS,
frequently applies to avoid a high loss with 90 declarer points,
called Schneider, or a maximum loss with 120 declarer points.

Therefore, once the contract of the declarer has been achieved,
we use the KBPS algorithm in opponent play with a 
scoring limit for Schneider/Schwarz.

\section{Experiments}

The Skat AI is written in C++, compiled with
\verb\gcc\ version 4.9.2 (optimization level \verb\-O2\).
Each websocket player client runs on 1 core of an
Intel Xeon Gold 6140 CPU @ 2.30GHz.
From the start to the very end of the game, the Skat AIs act as fully independent programs, that can play either on a server or replay human expert games. 

We determine the average of the game value according to the extended Seeger-(Fabian-)System, the international agreed DSKV standard for 
evaluating game play, set to series of 36 games. The score is based on the number of wins and losses of each player in the series, and the game value of the games being played. For a single game $g$, the outcome is $V(D,g)$, if the game is won for the declarer $D$ and $-2\cdot V(D,g)$, if it is lost. In a series of games $G = g_1,\ldots,g_X$ these values are added for each player, so that $V(A,G) = V(A,g_1) + \ldots + V(A,g_X)$. 
Then, the evaluation strength of Player $A$ wrt. $B$ and $C$ 
is $V(A,G) + 50 \cdot (\#wins(A,G) - \#losses(A,G)) + 40 \cdot (\#losses(B,G) + \#losses(C,G)).$ 

\subsection{Database Play}

The obtained results on 50,000 human expert trump games\footnote{For
scientific cross-comparison we provide the set of games as supplementary material
of this paper} are presented in 
Tables~\ref{tab:aibidaiputresults}-- Table~\ref{tab:hbidhputresults}. For the three valid combinations of AI/Human bidding/discarding/game announcement we separate between the play with and without the support of 
Paranoia search. In the columns we further partition the game outcomes with respect to the declarer in the $i)$ original Human game play, $ii)$ an open card solver (that we call Glassbox), and $iii)$ AI trick-taking selfplay. 

In Table~\ref{tab:aibidaiputresults} we see that with the support of the Paranoia search, the declarer is able to win
$42,228-41,765 = 463$  more than the AIs without Paranoia search
and far more than the Humans in their play 
$42,228-41,283 = 945$. This is a significant progress, given that the number of wins was already high. The number of games won 
(and the extended Seeger values) were higher than the ones 
obtained by the humans.

\begin{table}[t]
    \centering
\scriptsize    \begin{tabular}{c|ccc|r|r}
        & Human & Glassbox & AI & $+$Paranoia & $-$Paranoia \\ 
                & Wins & Wins & Wins & Opponents & Opponents \\ \hline
         $-$Paranoia
          &  false & false & false &  2,563 &  2,530  \\
         Declarer  
           &  false & false & true & 2,208 & 2,241  \\
           &  false & true & false &  231 & 226  \\
           &  false & true & true & 2,658 & 2,663  \\ 
           &  true & false & false & 3,438 & 3,407  \\
           &  true & false & true & 5,540 & 5,571  \\
           &  true & true & false & 975 &  970 \\
           &  true & true & true & 31,285 & 31,290  \\ \hline 
       Total +PO & 41,283 & 35,149 & 41,691 & 48,898 & -  \\ \hline
       Total -PO & 41,283 & 35,149 & 41,765 & - & 48,898   \\ \hline
         Total Score &   &  &  & 977.76  & 980.07   \\  \hline  
         Total Time &   &  &  &  37h:51m &  31h:01m  \\  \hline  \hline
         $+$Paranoia
           &  false & false & false &  2,460 &  2,458 \\
         Declarer  
           &  false & false & true &  2,315 & 2,313  \\
           &  false & true & false &  193 & 194  \\
           &  false & true & true &  2,696 & 2,695  \\ 
           &  true & false & false &  3,236 & 3,240  \\
           &  true & false & true &   5,742 & 5,738  \\
           &  true & true & false &  785 & 788 \\
           &  true & true & true &  31,475 & 31,472 \\ \hline 
       Total +PO & 41,283 & 35,149 & 42,228 & 48,898 & -  \\ \hline
       Total -PO & 41,283 & 35,149 & 42,218 & - & 48,898   \\ \hline
          Total Score &   &  &  & 990.25  & 991.79   \\ \hline 
          Total Time &   &  & & 43h:32m   & 42h:06m \\  \hline 
    \end{tabular}
    \caption{Skat AI Replaying 50,000 Human Trump Games with and without KBPS, using AI Bidding Game Selection and Skat Putting. Score is extended Seeger averaged for 36 games. Table split on whether or not KBPS is applied for the Declarer and the Opponents.  Total of games is smaller because of $1,102$ foldings (no bid).} 
    \label{tab:aibidaiputresults}
\end{table}

\begin{table}[t]
    \centering
\scriptsize    \begin{tabular}{c|ccc|r|r}
        & Human & Glassbox & AI &  $+$Paranoia & $-$Paranoia \\ 
        & Wins & Wins & Wins &  Opponents & Opponents \\ 
        \hline
        $-$Paranoia
           &  false & false & false &  3,854 & 3,822 \\
        Declarer
           &  false & false & true &  2,779 &  2,811 \\
           &  false & true & false &  198 &  195 \\
           &  false & true & true &  1,040 & 1,043 \\ 
           &  true & false & false &  2,137 & 2,110 \\
           &  true & false & true &  5,271 & 5,298 \\
           &  true & true & false &  965 &  953 \\
           &  true & true & true &  33,756 &  33,768 \\ \hline
          Total$+$PO & 42,129 & 35,959 & 42,846 & 50,000 & - \\ \hline 
          Total$-$PO & 42,129 & 35,959 & 42,920 & - & 50,000 \\ \hline 
          Total Score &   &  &  & 953.31 &  955.91   \\ \hline 
          Total Time &   &  &  & 34h:20m & 25h:31m     \\ \hline   \hline
         $+$Paranoia 
           &  false & false & false & 3,784 & 3755 \\
         Declarer
           &  false & false & true & 2,849 & 2,878 \\
           &  false & true & false & 184 & 128 \\
           &  false & true & true & 1,054 & 1,056 \\ 
           &  true & false & false & 1,998 & 1,973 \\
           &  true & false & true & 5,410 & 5,435 \\
           &  true & true & false & 761 & 752 \\
           &  true & true & true & 33,960 & 33,969 \\ \hline 
         Total + PO & 42,129 & 35,959 & 43,273 & 50,000 & - \\ \hline 
         Total - PO & 42,129 & 35,959 & 43,338 & - & 50,000  \\ \hline 
         Total Score &  &  &  & 963.35 & 965.53 \\ \hline 
         Total Time  &  &  &  & 47h:21m & 38h:05m \\ \hline 

    \end{tabular}
    \caption{Skat AI Replaying 50,000 Human Trump Games with and without KBPS using Human Bidding, Game Selection, and Skat Putting. Score is extended Seeger averaged for 36 games. Table split on whether or not KBPS is applied for the Declarer and the Opponents.} 
    \label{tab:hbidaiputresults}
\end{table}

\begin{table}[t]
    \centering
\scriptsize    \begin{tabular}{c|ccc|r|r}
        & Human & Glassbox & AI & +Paranoia & -Paraonoia \\ 
        & Wins & Wins & Wins & Opponents (PO)  & Opponents (PO) \\ 
        \hline
         $-$Paranoia
           &  false & false & false & 3,941  & 3,894 \\
         Declarer
           &  false & false & true & 2,519 & 2,566  \\
           &  false & true & false & 200  & 196 \\
           &  false & true & true & 1,211  & 1,215  \\ 
           &  true & false & false & 2,611  & 2,581   \\
           &  true & false & true & 5,502 & 5,532   \\
           &  true & true & false & 934  & 923  \\
           &  true & true & true &  33,082 & 33,093 \\ \hline  
         Total +PO & 42,129 & 35,959 & 42,314 & 50,000 & -  \\ \hline
        Total -PO & 42,129 & 35,959 &  42,406 & - & 50,000   \\ \hline
        Total Score &  &  &  & 935.23 & 938.41   \\ \hline 
        Total Time &   &  &  & 34h:10m & 25h:15m     \\ \hline 
           \hline
         $+$Paranoia 
           &  false & false & false & 3,853 & 3,808  \\
         Declarer
           &  false & false & true & 2,607 & 2,652  \\
           &  false & true & false & 187 & 185  \\
           &  false & true & true & 1,224 & 1,227 \\ 
           &  true & false & false & 2,452 & 2,415  \\
           &  true & false & true & 5,661 & 5,698  \\
           &  true & true & false & 732 & 724 \\
           &  true & true & true & 33,284 & 33,292  \\ \hline
Total +PO & 42,129 & 35,959 & 42,776 & 50,000 & -  \\ \hline
        Total -PO & 42,129 & 35,959 & 42,869 & - & 50,000   \\ \hline 
        Total Score &  &  &  & 947.41 &  950.39   \\ \hline 
        Total Time &  &  &  & 37h:39m & 46h:49m    \\ \hline 
    \end{tabular}
    \caption{Skat AI Replaying 50,000 Human trump games with and without KBPS using Human Bidding, and AI Skat Putting. Score is extended Seeger averaged for 36 games. Table split on whether or not KBPS is applied for the Declarer and the Opponents.} 
    \label{tab:hbidhputresults}
\end{table}

The AIs with KBPS show better winning ratios than the humans, and a significant positive effect 
on the playing strength in
extended Seeger score: for AI bidding and 
Skat putting almost $1,000$ points.
With up to $50$\% additional time, there is a
computational trade-off, but in
server play selecting the card to play
remains below $5$s.
In contrast
to Null games, automated Skat putting in trump
is worse to the Human one, and, therefore, subject to further research.
For AI bidding the total of wins/losses is not matching the total number of games, as some games might be folded.

We analyzed another set of over 75 thousand human expert games (different source, all kinds). We varied the card number $k$ to start invoking approximate KBPS at card
$k$ and KBPS at card $k+3$.
At $k=6$ we reached 1000.24 extended Seeger scoring points.
At $k=3$ we could slightly improve the value to 1001.39,
in a tradeoff of a slowdown (factor 2-3).

\subsection{Server Play Against Humans}

We also track games of human players challenging our AIs on the server and selected the last 100 series. 
As humans tend to fold 
when they are trailing behind, 
AI results look worse than they are\footnote{Unfortunately, unfinished 
series are not stored by the server. We could record the result
within the AIs, but we prefer the server evaluation, which we
prefer not to change.}. 
Still, even top Skat players frequently fail to win series, 
and the AIs scoring $1,342$ wins against $248$ losses with an
overall average of $977.5$ 
in the extended Seeger scoring system.

\section{Conclusion}

We have seen an improvement for knowledge inference in searching partial information games. The novelty is to include knowledge representation and reasoning into the 
backtrack partial-information 
game-tree search.
In contrast to \emph{perfect-information Monte-Carlo sampling} used by many AI card playing systems~\cite{GIB}, with a search for a sampled set of worlds, the KBPS
search algorithm operates against all possible worlds in one search tree, avoiding the fusion of different card suggestion and resulting in a single card recommendation. It progresses the knowledge in the search tree in an efficient manner, resulting in an optimal search algorithm is fast enough to be applied in early stages of the game even after a few cards have been played and especially for declarer play, leads to card suggestions that even experienced humans often do not see. If the analysis succeeds, 
this \emph{killer card} is forced. If not, other
card recommendations like expert rules or end game play apply.
Although exemplified for \emph{Skat}, the contribution is
general to work for other multi-player card games like \emph{Spades}, \emph{Hearts}, \emph{Tarot}, \emph{Marias}, \emph{Ulti},
or \emph{Bridge}, and likely to other domains. 
The results in increased playing strength especially for the declarer are unexpectedly promising. Even top players are astonished losing against the automated play to victory, once it had been found. 
More research is needed to increase the knowledge 
in the search tree, especially for opponent play.

We envision using (approximate) KBPS to 
decide, if a \emph{Grand} should be played, either by 
precomputation efforts and/or by dropping 
lost cards from the analysis. There are 
further research avenues towards approximate KBPS search, as the virtue of good card play is a card selection to win not against all but most possible distributions. 


\medskip

\paragraph*{Acknowledgements} Thanks to Rainer Gößl, a world-class Skat player, who helped pushing this project forward with his Skat expertise.

\begin{figure}[t]
\begin{scriptsize}
\begin{lstlisting}
AND(avail) 
  while (avail) 
    index = select(avail); bit = (1<<index);
    (h0,h2,o,as,ms) = 
         (h[0],h[2],pool,declarerorskat,partnerorskat);
    if (c = firstcardontable(i)) 
      if (trump & (1 << c)) 
        if (|trump & bit| == 0) 
          partnerorskat |= trump & pool; 
          if (|partnerorskat| > 2 + r2 - rearhand)   
            h[2] = h2; partnerorskat = ms;
            avail &= ~bit; continue;
          pool &= ~partnerorskat; 
          h[2] |= noskat & partnerorskat; 
          if (|h[2]| > r2 - h) 
            (h[2],pool,partnerorskat) = (h2,o,ms); 
            avail &= ~bit; continue;
          partnerorskat &= ~h[2];
      else 
        if (|suit(c) & bit| == 0)  
          partnerorskatat |= suit(c) & o;
          if (|partnerorskat| > 2 + r2 - rearhand) 
            (h[2],t,partnerorskat) = (h2,o,ms); 
            avail &= ~bit; continue;
          pool &= ~partnerorskat; 
          h[2] |= noskat & partnerorskat; 
          if (|h[2]| > r2 - rearhand) 
            (h[2],pool,partnerorskat) = (h2,o,ms); 
            avail &= ~bit; continue;
          partnerorskat &= ~h[2]; 
    if (|h[0]|bit| > r0)   
     (h[2],partnerorskat,pool) = (h2,ms,o); 
     avail &= ~bit; 
     continue; 
    p = played; pool &= ~bit;h[0] &= ~bit;h[2] &= ~bit; 
    declarerorskat &= ~bit; partnerorskat &= ~bit;
    played |= bit; i[0] = index; r = -1; 
    if (endoftrick) 
      r0--; r2--; w = winner(1,2,0);
      score = value(i); ap = aspts; gp = gspts;
      gspts += w ? score: 0; aspts += !w ? score: 0;
      t1 = i[1], t2 = i[2];
      if (|played| == 30) aspts += value(~played); 
      i[0] = i[1] = i[2] = -1;
      r = (gspts >= 120-limit) ? 0 : (aspts > limit) ? 1:
        (w==0)? AND((pool|declarerorskat|h[0]) & ~h[2]):
        (w==1)? OR1(h[1]): 
        OR2((o|partnerorskat|h[2]) & ~h[0]);
      i[1] = t1; i[2] = t2;
      gspts = gp; aspts = ap; r0++; r2++; 
    else r = OR1(feasible(h[1],i));
    (h[0],h[2],played,declarerorskat,partnerorskat,pool) =
       (h0,h2,p,as,ms,o); 
    avail &= ~bit; i[0] = -1;
    if (r == 1) return 1;  
  return 0;
\end{lstlisting}
\end{scriptsize}
\caption{KBPS search algorithm for the opponent $P_1$ at a declarer search AND node of the serach tree. }\label{triple}
\end{figure}


\bibliographystyle{IEEEtran}
\bibliography{para.bib}

\end{document}